\documentclass[conference]{IEEEtran}
\IEEEoverridecommandlockouts
\usepackage{cite}
\usepackage{amsmath,amssymb,amsfonts}
\usepackage{algorithmic}
\usepackage{graphicx}
\usepackage{textcomp}
\usepackage{xcolor}
\usepackage{enumitem}
\usepackage{array}
\usepackage{placeins}

\usepackage{lipsum,booktabs}

\usepackage{booktabs} 
\usepackage{float} 
\usepackage{graphicx} 
\usepackage{arydshln} 

\usepackage{hyperref} 
\usepackage{url} 
\hypersetup{
    colorlinks=true,
    linkcolor=blue,
    urlcolor=blue,
    citecolor=blue
}

\def\BibTeX{{\rm B\kern-.05em{\sc i\kern-.025em b}\kern-.08em
    T\kern-.1667em\lower.7ex\hbox{E}\kern-.125emX}}
    
\begin{document}

\pagestyle{plain}

\title{Enhancing Diffusion Face Generation with Contrastive Embeddings and SegFormer Guidance}

\author{
\IEEEauthorblockN{Dhruvraj Singh Rawat\IEEEauthorrefmark{1}, Enggen Sherpa\IEEEauthorrefmark{1}, Rishikesan Kirupanantha\IEEEauthorrefmark{1}, Tin Hoang\IEEEauthorrefmark{1}\IEEEauthorrefmark{2}}
\IEEEauthorblockA{University of Surrey, UK}
\IEEEauthorblockA{\{dr00732, es02280, rk01337, th01167\}@surrey.ac.uk}
\IEEEauthorblockA{\IEEEauthorrefmark{1}Authors' names are listed in alphabetical order, \IEEEauthorrefmark{2}Team lead}
}

\maketitle

\begin{abstract}
We present a benchmark of diffusion models for human face generation on a small-scale CelebAMask-HQ dataset, evaluating both unconditional and conditional pipelines. Our study compares UNet and DiT architectures for unconditional generation and explores LoRA-based fine-tuning of pretrained Stable Diffusion models as a separate experiment. Building on the multi-conditioning approach of Giambi and Lisanti \cite{giambi2023conditioning}, which uses both attribute vectors and segmentation masks, our main contribution is the integration of an InfoNCE loss for attribute embedding and the adoption of a SegFormer-based segmentation encoder. These enhancements improve the semantic alignment and controllability of attribute-guided synthesis. Our results highlight the effectiveness of contrastive embedding learning and advanced segmentation encoding for controlled face generation in limited data settings.
\footnote{Source code available at \url{https://github.com/Tin-Hoang/humanfaces-diffusion-generation}}
\end{abstract}


\section{Introduction}
Diffusion models have become a leading approach for high-quality image generation, offering stable training and diverse outputs compared to GANs~\cite{dhariwal2021diffusion,ho2020denoising}. They are particularly well-suited for human face synthesis, which requires fine-grained control over identity and attributes~\cite{giambi2023conditioning}. CelebAMask-HQ \cite{lee2020maskgan} provides a suitable benchmark with high-resolution images and detailed attribute and segmentation annotations. In this work, we benchmark unconditional and conditional diffusion pipelines for face generation, and introduce improvements to attribute conditioning by leveraging InfoNCE loss and a SegFormer-based segmentation encoder.

\section{Related Works}
Diffusion models have become state-of-the-art for high-fidelity image synthesis, including human faces. Unconditional models like DDPM~\cite{ho2020denoising} and its UNet-based variants~\cite{dhariwal2021diffusion} learn to model data distributions without external signals, while transformer-based backbones (e.g., DiT~\cite{Peebles2022DiT}) improve long-range dependency modeling. Parameter-efficient adaptation methods such as LoRA~\cite{hu2022lora} enable fast fine-tuning of large diffusion models.

Conditional diffusion extends this framework by injecting auxiliary information (e.g., class labels, attributes) into the generative process~\cite{zhan2024conditional}. Latent diffusion~\cite{rombach2022high} further reduces computational cost by operating in a compressed latent space, with conditioning typically achieved via cross-attention to attribute or segmentation embeddings~\cite{zhang2023adding}. Contrastive objectives like InfoNCE~\cite{oord2018representation} help ensure attribute embeddings are semantically meaningful.

For face generation, Giambi \textit{et al.}~\cite{giambi2023conditioning} demonstrated that conditioning diffusion models on both attribute vectors and semantic masks yields precise, high-quality facial synthesis. Our approach builds on this baseline, introducing InfoNCE-based attribute embedding and replacing the ResNet-18 mask encoder with a SegFormer~\cite{xie2021segformer} backbone for improved spatial conditioning.

\section{Methodology}
\begin{figure*}[t]
\centering
\includegraphics[width=\textwidth]{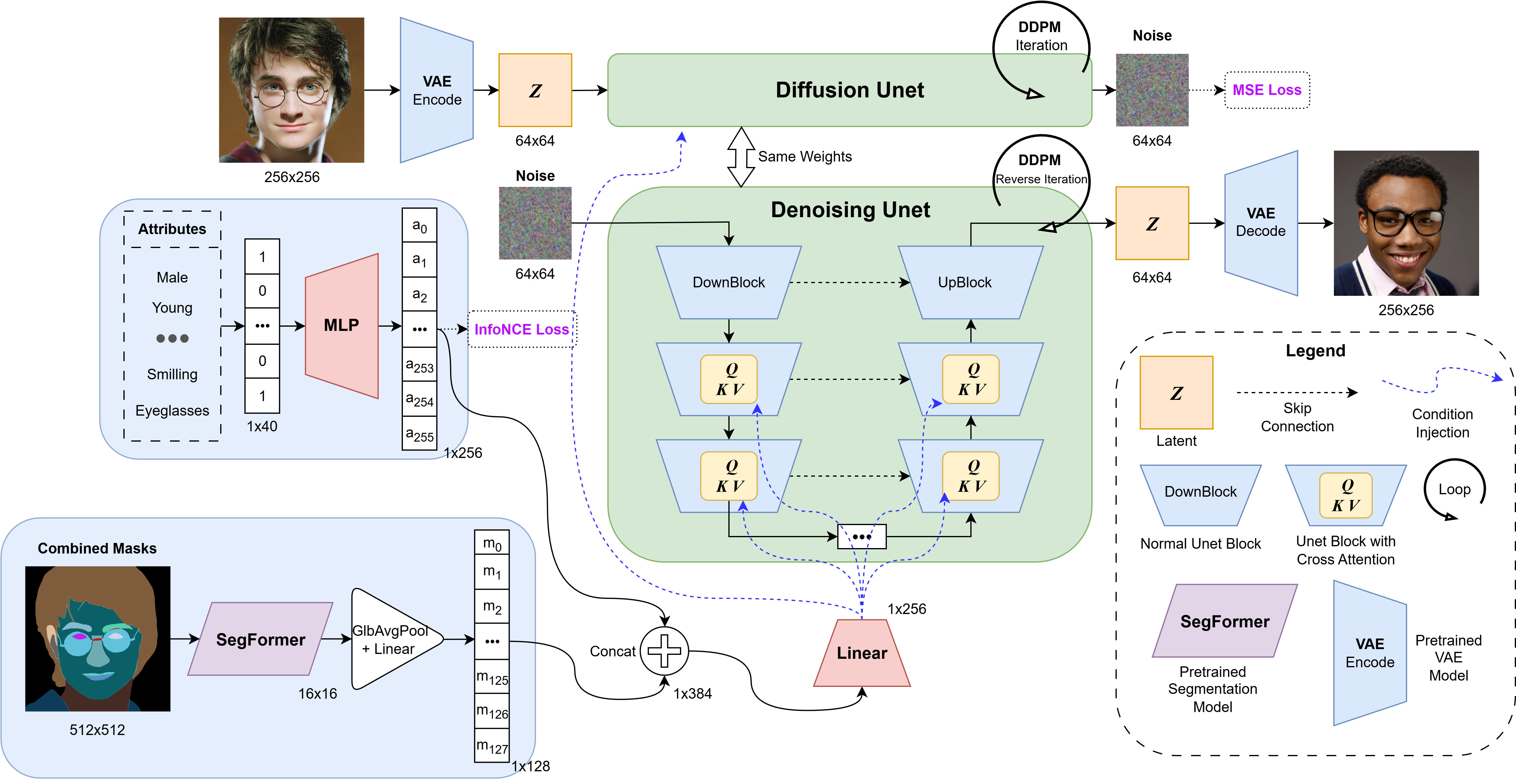}
\caption{Overview of the attribute-conditioned diffusion pipeline. Attribute embeddings and segmentation mask encodings are incorporated to guide image synthesis. In attribute-only experiments, the attribute embedding is fed directly into the U-Net, omitting both the concatenation with spatial features and the linear transformation module.}
\label{fig:attributes_pipeline}
\end{figure*}

\subsection{Unconditional Diffusion with Unet}

We implemented several UNet architectures using the \texttt{UNet2DModel} from the Hugging Face \texttt{diffusers} library. Our architecture variants explore different design choices. Details as in \ref{appendix:unet_variants}. To improve stability and sample quality, we apply an Exponential Moving Average (EMA) strategy on model weights.

\subsection{DiT Backbone for Diffusion}

We implemented the Diffusion Transformer (DiT) \cite{Peebles2022DiT} as a transformer-based alternative to UNet for diffusion models. We experimented with two variants (see Appendix \ref{appendix:dit_arch}), both trained with DDPM and EMA for stability. Originally designed for class-conditional generation, we adapt DiT for unconditional generation by using dummy labels. This allows a direct comparison with UNet-based approaches.

\subsection{Stable Diffusion Fine-Tuning with LoRA}
We fine-tuned Stable Diffusion v2 using Low-Rank Adaptation (LoRA) via the Hugging Face \texttt{PEFT} library to adapt it for high-fidelity face generation. LoRA adds rank-decomposed matrices to UNet attention layers, enabling efficient adaptation in the latent space with a pretrained VQ-VAE encoder, compressing $512 \times 512$ images into $64 \times 64 \times 4$ latents~\cite{rombach2022high}.

\subsection{Human Faces Attributes Conditioning Generation}
Our approach for attribute-conditioned human face generation builds on the baseline by Giambi \textit{et al.}~\cite{giambi2023conditioning}, which conditions diffusion models using attributes and semantic masks. We introduce two key differences: (1) InfoNCE loss~\cite{oord2018representation} for training the attribute embedder to enhance embedding discriminability, and (2) a pretrained SegFormer~\cite{xie2021segformer} for segmentation embedding instead of ResNet18.

The pipeline (see Fig. \ref{fig:attributes_pipeline}) leverages a conditional latent diffusion framework optimized for CelebAMask-HQ, with three main components:
\begin{itemize}[leftmargin=*]
    \item \textbf{Latent Space Encoding:} A pre-trained VQ-VAE~\cite{rombach2022high} compresses $256 \times 256 \times 3$ images into a $64 \times 64$ latent space.
    \item \textbf{Conditional Diffusion Model:} A \texttt{UNet2DConditionModel} denoises latents, conditioned on 256D embeddings via cross-attention, using the DDPM framework~\cite{ho2020denoising}.
    \item \textbf{Attribute Embedder:} A Multi-Layer Perceptron maps 40D multi-hot attribute vectors to 256D embeddings. This module is optimized by InfoNCE loss.
    \item \textbf{Segmentation Embedder:} Pretrained SegFormer~\cite{xie2021segformer} processes combined segmentation masks into spatial embeddings, and uses Global Average Pooling with Linear layer to map it to 128D embeddings.
\end{itemize}

For attributes-only experiments, attribute embeddings directly condition the UNet. Joint attribute and segmentation embeddings are concatenated and projected ($512 \rightarrow 256$) for precise spatial and attribute control.

\subsection{InfoNCE Loss for Attribute Embedding}
The InfoNCE (Noise-Contrastive Estimation) loss is employed to train the attribute embedder, ensuring that 256-dimensional embeddings capture semantic relationships among multi-hot attribute vectors \cite{oord2018representation}.

For a batch of embeddings \( z \) (shape \([B, 256]\)), where \(B\) is batch size) and multi-hot attribute vectors \( a \) (shape \([B, 40]\)), the InfoNCE loss is computed as follows:

1. \textbf{Attribute Similarity:}
   - Cosine similarity is calculated between all pairs of attribute vectors, forming a similarity matrix \([B, B]\). Pairs with similarity above a threshold (e.g., 0.8) are marked as positive, indicating high attribute overlap (e.g., "smiling + glasses" vs. "smiling").

2. \textbf{Embedding Similarity:}
   - Cosine similarity is computed between all pairs of embeddings, scaled by a temperature parameter \( \tau \) (default 0.07), yielding an embedding similarity matrix \([B, B]\).

3. \textbf{Positive and Negative Pairs:}
   - A positive mask is generated from the attribute similarity matrix, assigning 1 to positive pairs and 0 to others.

4. \textbf{Loss Formulation:}
   - The InfoNCE loss encourages embeddings of positive pairs to be close while distancing negative pairs, defined as:
     \[
     L_{attr} = -\frac{1}{B} \sum_{i=1}^B \log \left( \frac{\sum_{j \in \text{pos}} \exp(\text{sim}(z_i, z_j)/\tau)}{\sum_{k=1}^B \exp(\text{sim}(z_i, z_k)/\tau)} \right)
     \]
     where \( \text{sim}(\cdot, \cdot) \) is cosine similarity, and \( \text{pos} \) denotes indices of positive pairs.

\section{Experimental Results}

\subsection{Unconditional Diffusion}
\textbf{Dataset} We evaluated unconditional diffusion using two CelebA-HQ configurations: 
\begin{itemize}[nosep,leftmargin=*]
    \item \textbf{Small-scale:} 2.7k training images (256$\times$256→128$\times$128) + 300 test images
    \item \textbf{Full-scale:} 27k images (1024$\times$1024→128$\times$128) with matched test set
\end{itemize}
\textbf{Models.} Three UNet variants (\texttt{r3/r5/r6}) (see \ref{appendix:unet_variants}) were implemented, varying in depth (4-6 blocks), attention placement, and channel width. All models used DDPM at 128$\times$128 resolution with EMA stabilization, trained with learning rates 1e-5–5e-4 and warmup schedules.

\begin{table}[ht]
\centering
\caption{Experimental Results of Unconditional Models}
\label{tab:uncond_models}
\small
\begin{tabular}{l c c c c}
\toprule
\textbf{Model} & \textbf{Learning Rate / Warmup} & \textbf{EMA} & \textbf{FID} $\downarrow$ \\
\midrule

\multicolumn{5}{l}{\textit{Different UNet Architectures Experiments}} \\
unet\_r3 & 2e-4 / 3000 & Yes & \textbf {72.6211} \\
unet\_r5 & 2e-4 / 3000 & Yes & 76.0807 \\
unet\_r6 & 2e-4 / 1500 & Yes & 73.3532 \\

\midrule
\multicolumn{5}{l}{\textit{Learning Rate Experiments (R5)}} \\
unet\_r5 & 1e-5 / default & Yes & 99.4773 \\
unet\_r5 & 5e-4 / default & Yes & 75.3131 \\

\midrule
\multicolumn{5}{l}{\textit{Warmup Strategy Experiments (R5)}} \\
unet\_r5 & 2e-4 / 3000 & Yes & 76.4401 \\
unet\_r5 & 2e-4 / 10000 & Yes & 73.9984 \\

\midrule
\multicolumn{5}{l}{\textit{EMA Ablation (R5)}} \\
unet\_r5 & 2e-4 / 500 & \textbf{No} & 92.8969 \\

\midrule
\multicolumn{5}{l}{\textit{DiT Architectures Experiments}} \\
dit\_large\_2.7k            & 2e-4 / 3000 & Yes  & 89.90 \\
dit\_small\_2.7k            & 2e-4 / 3000 & Yes  & 94.00 \\

\bottomrule
\end{tabular}
\end{table}

Our experiments reveal critical insights into UNet design and training strategies for unconditional face generation:

\subsubsection{UNet Architecture:} Deeper variants (\texttt{unet\_r3/r5}) with mid-level attention achieved superior FID (72.62--76.08) compared to shallower designs, validating the importance of strategic attention placement (Table~\ref{tab:uncond_models}).
\subsubsection{Training Dynamics:} Optimal performance (FID 73.99) used 2e-4 LR with 10k warmup steps-excessive rates (5e-4) caused instability, while low rates (1e-5) led to underfitting.
\subsubsection{EMA Impact:} Weight averaging reduced FID by $\sim$19 points (92.89→73.99), confirming its necessity for stable convergence.

Full results across architectural and training configurations are detailed in Table~\ref{tab:uncond_models}. The findings underscore that architectural depth combined with careful training stabilization enables high-quality face synthesis.

\subsubsection{DiT Experiment}
DiT-Large outperforms DiT-Small on the 2.7k dataset (FID 89.90 vs. 94.00), indicating that increased model capacity improves generalization even with limited training data.

\subsection{Conditional Diffusion via Attributes}

\begin{table*}[t]
\centering
\caption{Experimental Results of Attributes Conditioning Models. All entries pertain to attribute-only models, except the final one, which incorporates both attributes and segmentation encoding.}
\label{tab:attributes-fid}
\begin{tabular}{lcccccc}
\toprule
\textbf{Model} & \multicolumn{1}{c}{\textbf{\#Params}} & \multicolumn{1}{l}{\textbf{\#DDPM steps}} & \multicolumn{1}{l}{\textbf{\#Train Samples}} & \multicolumn{1}{l}{\textbf{VAE type}} & \multicolumn{1}{l}{\textbf{InfoNCE}} & \multicolumn{1}{l}{\textbf{FID} $\downarrow$} \\ \midrule
\multicolumn{7}{c}{\textbf{Different Conditional UNet Architectures Experiments}} \\ \hdashline[2pt/3pt]
\textbf{LC\_UNet\_Base} & 140.45 M & 1000 & 2700 & VQ-VAE & Yes & 75.4572 \\
\textbf{LC\_UNet\_3} & 104.95 M & 1000 & 2700 & VQ-VAE & Yes & \textbf{70.9824} \\
\textbf{LC\_UNet\_5} & 168.04 M & 1000 & 2700 & VQ-VAE & Yes & 77.6988 \\
\textbf{LC\_UNet\_6} & 116.84 M & 1000 & 2700 & VQ-VAE & Yes & 74.4457 \\ \midrule
\multicolumn{7}{c}{\textbf{DDPM Steps Experiments}} \\ \hdashline[2pt/3pt]
\textbf{LC\_UNet\_Base} & 140.45 M & 1000 & 2700 & VQ-VAE & Yes & \textbf{75.4572} \\
\textbf{LC\_UNet\_Base} & 140.45 M & 2000 & 2700 & VQ-VAE & Yes & 77.4334 \\ \midrule
\multicolumn{7}{c}{\textbf{VAE Type Experiments}} \\ \hdashline[2pt/3pt]
\textbf{LC\_UNet\_3} & 104.95 M & 1000 & 2700 & VQ-VAE & Yes & \textbf{70.9824} \\
\textbf{LC\_UNet\_3} & 104.95 M & 1000 & 2700 & VAE & Yes & 70.9829 \\ \midrule
\multicolumn{7}{c}{\textbf{Number of Train Samples Experiments}} \\ \hdashline[2pt/3pt]
\textbf{LC\_UNet\_3} & 104.95 M & 1000 & 2700 & VQ-VAE & Yes & \textbf{70.9824} \\
\textbf{LC\_UNet\_3} & 104.95 M & 1000 & 27000 (x10 times) & VQ-VAE & Yes & 74.7878 \\ \midrule
\multicolumn{7}{c}{\textbf{InfoNCE Ablation Study}} \\ \hdashline[2pt/3pt]
\textbf{LC\_UNet\_3} & 104.95 M & 1000 & 2700 & VQ-VAE & No & 74.0719 \\
\textbf{LC\_UNet\_3} & 104.95 M & 1000 & 2700 & VQ-VAE & Yes & \textbf{70.9824} \\ \bottomrule
\multicolumn{7}{c}{\textbf{Segmentation Mask Experiments}} \\ \hdashline[2pt/3pt]
\textbf{LC\_UNet\_3} (Attributes Only) & 104.95 M & 1000 & 2700 & VQ-VAE & Yes & 70.9824 \\
\textbf{LC\_UNet\_3} (Attributes + Segmentation) & 104.95 M + 3.73 M & 1000 & 2700 & VQ-VAE & Yes & \textbf{63.85} \\ \bottomrule
\end{tabular}
\end{table*}

\textbf{Dataset.} We used the CelebAMask-HQ dataset, containing 30,000 high-resolution face images with 40 binary attribute annotations (e.g., ``smiling,'' ``eyeglasses''). Most experiments used 2,700 images for training and 300 for testing, except the training sample size experiment, which used 27,000 training images.

\textbf{Models.} We experimented with four different Conditional Unet architectures training from scratch, see \ref{appendix:lc_unet}. The VAE and SegFormer were pre-trained and frozen during the training.

\textbf{Metrics.} Fréchet Inception Distance (FID) was computed between 300 real test images and images generated using attribute labels from the same test set.

\subsubsection{Attributes Conditioning}
Our conditional diffusion pipeline employs a latent-space UNet (\texttt{UNet2DConditionModel}) conditioned on attribute embeddings from 40-dimensional multi-hot vectors. A novel InfoNCE loss \cite{oord2018representation} enhances embedding discriminability, as detailed below. Table~\ref{tab:attributes-fid} presents results, with \texttt{LC\_Unet\_3} achieving the best FID of 70.9824 using 104.95M parameters, 1000 DDPM steps, and a VQ-VAE \cite{rombach2022high}.

\subsubsection{Different Conditional UNet Variants}
We evaluate multiple UNet architectures (Table~\ref{tab:attributes-fid}). \texttt{LC\_Unet\_3} (5 resolution blocks, selective cross-attention at 32x32$\to$16x16) outperforms deeper models like \texttt{LC\_Unet\_5} (168.04M parameters, FID 77.6988) and \texttt{LC\_Unet\_6} (116.84M parameters, FID 74.4457). The baseline \texttt{LC\_Unet\_Base} (140.45M parameters, FID 75.4572) is competitive, but \texttt{LC\_Unet\_3}'s efficient design optimizes attribute-guided generation.

\subsubsection{DDPM Steps Experiments}
Increasing DDPM steps from 1000 to 2000 for \texttt{LC\_Unet\_Base} raises FID from 75.4572 to 77.4334, indicating that 1000 steps suffice for convergence, balancing quality and efficiency.

\subsubsection{VAE Type Experiments}
We compare two autoencoders: a VAE from Stable Diffusion v1-5 \cite{stablediffusionv15} and a VQ-VAE from CompVis/ldm-celebahq-256 \cite{CompVis2022}. For \texttt{LC\_Unet\_3}, VQ-VAE (FID 70.9824) and VAE (FID 70.9829) yield nearly identical results. VQ-VAE’s 4x downsampling preserves more detail than VAE’s 8x, but the negligible FID gap suggests robustness to autoencoder type.

\subsubsection{Number of Training Samples}
Training \texttt{LC\_Unet\_3} with 27,000 samples increases FID to 74.7878 compared to 70.9824 with 2,700 samples\footnote{Training iterations kept constant between small/large datasets to control for learning dynamics}. This indicates that a smaller, curated dataset is sufficient for effective attribute conditioning, likely due to cleaner attribute annotations.

\subsubsection{InfoNCE Loss Ablation Study}
Our key contribution is the use of InfoNCE loss \cite{oord2018representation} to train the attribute embedder, ensuring embeddings capture semantic relationships. Ablating InfoNCE in \texttt{LC\_Unet\_3} raises FID from 70.9824 to 74.0719, demonstrating that InfoNCE significantly improves attribute-specific generation by aligning embeddings with attribute similarities, enhancing the UNet’s conditioning precision.

\subsubsection{Segmentation Mask Experiments}
To enhance attribute-conditioned face generation, we evaluated the impact of incorporating segmentation masks alongside attribute vectors in \texttt{LC\_UNet\_3}, as shown in Table \ref{tab:attributes-fid}. Segmentation masks provide spatial guidance, potentially improving the structural fidelity of generated faces.
Adding segmentation masks reduced the FID to 63.85, a significant improvement. This result indicates that segmentation masks enhance the model's ability to capture fine-grained facial structures, leading to higher-quality generated images. The combination of attribute conditioning and spatial guidance proves highly effective for human face synthesis.

\subsection{Conditional and Unconditional LoRA Adaptation of Stable Diffusion}
\textbf{Setup} We evaluated two fine-tuning pipelines (see Appendix~\ref{appendix:lora_stable_diffusion} for details of both setups). In this main report, we present Setup 1 only: Full-featured with dynamic sampling, text conditioning, EMA, and FID tracking. All models were trained on 27k CelebA-HQ images resized to $512 \times 512$. FID scores were computed on a held-out test set from our data split. Sampling was performed using 50 and 150 inference steps to compare generation quality across settings.

\begin{table}[h]
\centering
\caption{FID ($\downarrow$) scores on our test set for LoRA fine-tuned Stable Diffusion v2 (using Setup 1).}
\label{tab:lora-fid}
\begin{tabular}{lccc}
\toprule
\textbf{Inference Steps} & \textbf{Condition} & \textbf{Version} & \textbf{FID} $\downarrow$ \\
\midrule
50   & Unconditional     & No Tuning & 297.098 \\
150  & Unconditional     & No Tuning & 315.0361 \\
50  & Conditional     & No Tuning & 114.729 \\
150  & Conditional     & No Tuning & 120.4432 \\
\midrule
50   & Conditional     & LoRA & 67.515 \\
150  & Conditional     & LoRA & 65.305 \\
50   & Unconditional   & LoRA & 91.315 \\
150  & Unconditional   & LoRA & 119.082 \\
\bottomrule
\end{tabular}
\end{table}

\textbf{Key Observations}
As shown in Table \ref{tab:lora-fid}, fine-tuning Stable Diffusion v2 using LoRA has significantly improved FID scores, particularly for conditional models. Version 1 achieved the best FID of 65.305 at 150 inference steps. Conditional models consistently outperform their unconditional counterparts across all versions, aligning with the fact that the baseline Stable Diffusion model was originally trained on conditional datasets. This alignment likely contributes to the stronger performance of conditional fine-tuned models. Additionally, increasing inference steps improves conditional generation quality but degrades unconditional performance.

\section{Conclusion}
We presented a comprehensive study of human face generation with diffusion models, benchmarking both unconditional and conditional pipelines. For unconditional generation, our experiments demonstrated that carefully designed UNet architectures with mid-level attention and EMA stabilization achieve strong performance (best FID 72.62), while transformer-based DiT models, though promising, underperformed at comparable scales (best FID 89.90)~\cite{dhariwal2021diffusion,Peebles2022DiT}. We also showed that LoRA fine-tuning of Stable Diffusion v2 enables efficient adaptation to face domains, with conditional LoRA models achieving superior FID (65.31) over unconditional variants~\cite{rombach2022high}.

Building on the attribute- and segmentation-conditioned diffusion framework introduced by Giambi and Lisanti~\cite{giambi2023conditioning}, our main novelty is the integration of an InfoNCE contrastive loss for training attribute embeddings and the use of a SegFormer-based segmentation encoder~\cite{xie2021segformer}. This combination significantly improved attribute alignment and spatial control, reducing FID from 74.07 to 70.98 when incorporating InfoNCE loss. Our results validate that contrastive embedding learning and advanced segmentation encoding enhance the fidelity and controllability of attribute-guided face synthesis.

\clearpage 
\onecolumn 

\appendix 
\section{Appendix}

\subsection{Different Unconditional UNet Architectures} \label{appendix:unet_variants}

Our architecture variants explore different design choices:
\begin{itemize}
    \item Depth: from 4 to 6 down/up sampling blocks.
    \item Attention usage: some blocks include spatial self-attention (\texttt{AttnDownBlock2D}, \texttt{AttnUpBlock2D}).
    \item Channel width: starting from 128 and scaling up to 1024.
\end{itemize}

\begin{table*}[ht]
\centering
\caption{Comparison of Unconditional UNet Architectures (Variants r3 to r6)}
\label{tab:unet_r3_r6}
\begin{tabular}{@{}lccclcl@{}}
\toprule
\textbf{Model Name} & \textbf{Input Size} & \textbf{\# Blocks} & \textbf{Down Blocks}         & \textbf{Up Blocks}           & \textbf{Block Out Channels}         \\
\midrule
\texttt{unet\_r3} & 128$\times$128 & 6 & D, D, CA, D, CA, D         & U, CA, U, U, CA, U           & 128, 256, 256, 512, 512, 1024 \\

\texttt{unet\_r5} & 128$\times$128 & 6 & D, D, D, D, CA, D          & U, CA, U, U, U, U            & 128, 128, 256, 256, 512, 512   \\
\texttt{unet\_r6} & 128$\times$128 & 4 & D, CA, CA, D               & U, CA, CA, U                 & 160, 320, 640, 640             \\
\bottomrule
\end{tabular}
\end{table*}

\begin{figure}[h]
\centering
\includegraphics[width=0.5\linewidth]{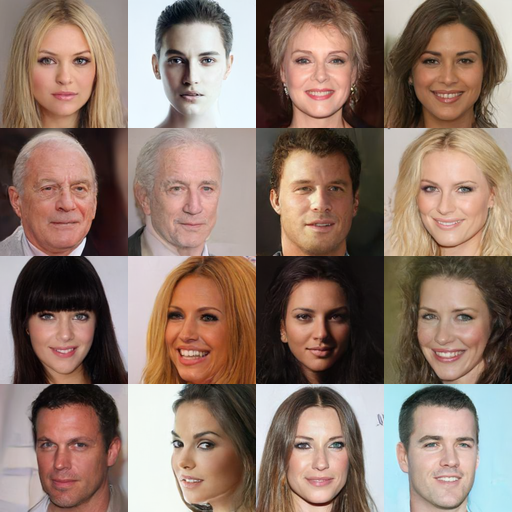}
\caption{Generated faces from the \texttt{unet\_r5} model, trained from scratch on 2,700 CelebA-HQ images provided by TA.}
\label{fig:unet_r5_visualization}
\end{figure}

\subsection{Different DiT Architectures} \label{appendix:dit_arch}

\begin{itemize}
    \item \textbf{DiT-Small}: A lightweight model with 6 transformer layers, 8 attention heads, and 512-dimensional hidden size. It uses 4×4 patching for 128×128 images and applies a dropout rate of 0.1.
    
    \item \textbf{DiT-Large}: A deeper model with 12 transformer layers, 12 attention heads, and a 768-dimensional hidden size. It shares the same patching and normalization strategy as DiT-Small, with a stronger dropout of 0.2.
\end{itemize}

\begin{table*}[ht]
\centering
\caption{Comparison of DiT Backbone Architectures (DiT-Small and DiT-Large)}
\label{tab:dit-architectures}
\begin{tabular}{lcccccc}
\toprule
\textbf{Model Name} & \textbf{Input Size} & \textbf{\# Layers} & \textbf{Patch Size} & \textbf{Hidden Dim} & \textbf{\# Heads} & \textbf{Dropout} \\
\midrule
\texttt{dit\_small} & 128×128 & 6  & 4×4 & 512 & 8  & 0.1 \\
\texttt{dit\_large} & 128×128 & 12 & 4×4 & 768 & 12 & 0.2 \\
\bottomrule
\end{tabular}
\end{table*}

\textbf{Setup} We evaluated DiT-Small (6 layers, 512 hidden dim) and DiT-Large (12 layers, 768 hidden dim) in CelebA-HQ at $128 \times 128$ resolution using 1\,000 DDPM timesteps. Models are trained under two data regimes: 2.7k and 27k images. We assess performance with and without input normalization. The "no\_norm" runs omits \texttt{ transforms. Normalize ([0.5], [0.5])}, keeping pixel values in the range \([0,1]\).

\begin{table}[h]
\centering
\caption{FID ($\downarrow$) for DiT variants.}
\label{tab:dit-fid}
\begin{tabular}{lccc}
\toprule
\textbf{Run} & \textbf{Variant} & \textbf{Images} & \textbf{FID} $\downarrow$ \\
\midrule
\texttt{dit\_large\_2.7k}            & Large & 2.7k  & 89.90 \\
\texttt{dit\_small\_2.7k}            & Small & 2.7k  & 94.00 \\
\texttt{dit\_large\_27k}             & Large & 27k   & 99.47 \\
\texttt{dit\_large\_27k\_no\_norm}   & Large & 27k   & 93.92 \\
\texttt{dit\_small\_27k}             & Small & 27k   & 91.20 \\
\texttt{dit\_small\_27k\_no\_norm}   & Small & 27k   & 92.04 \\
\bottomrule
\end{tabular}
\end{table}

\textbf{Key Observations}
\begin{itemize}[nosep,leftmargin=*]
    \item DiT-Large outperforms DiT-Small on the 2.7k dataset (FID 89.90 vs. 94.00), indicating that increased model capacity improves generalization even with limited training data.

    \item Scaling DiT-Small to 27k images leads to improved performance (94.00 → 91.20), confirming the expected benefit of more data for smaller architectures.

    \item In contrast, DiT-Large performs worse when trained on 27k images compared to 2.7k (FID 99.47 vs. 89.90), suggesting increased sensitivity to training dynamics or overfitting when both model depth and data scale increase.

    \item Removing pixel normalization improves FID for both model sizes at the 27k scale (e.g., DiT-Large: 99.47 → 93.92), but results in cloudy and low-contrast images. This indicates that while normalization impacts feature statistics used in FID computation, it does not necessarily align with perceptual quality. The lower FID may result from reduced sample diversity, causing tighter clustering in the InceptionV3 feature space. These findings emphasize the importance of evaluating sample quality both quantitatively and visually, especially when image preprocessing alters the distribution characteristics.
\end{itemize}

\subsection{Different Attributes Conditional Unet Architectures} \label{appendix:lc_unet}

The comparison of Latent Conditional UNet architectures in Table \ref{tab:unet_vertical_comparison} evaluates the impact of block configurations and cross-attention placement on performance in conditional diffusion tasks. The table details the number of blocks, channel sizes, down-block types, and parameter counts, providing insight into architectural efficiency. LC\_UNet\_3 likely achieves superior performance due to its cross-attention at the second down block (32$\times$32 to 16$\times$16), effectively integrating conditioning information at an intermediate resolution to capture both global and local features. With fewer blocks (5) and parameters (104.95M), it may be better generalized, avoiding overfitting compared to models like LC\_UNet\_5 (168.04M parameters).

\subsection{InfoNCE loss Impact Visualisation} \label{appendix:infonce_viz}
\begin{figure}[ht]
\centering
\includegraphics[width=0.6\linewidth]{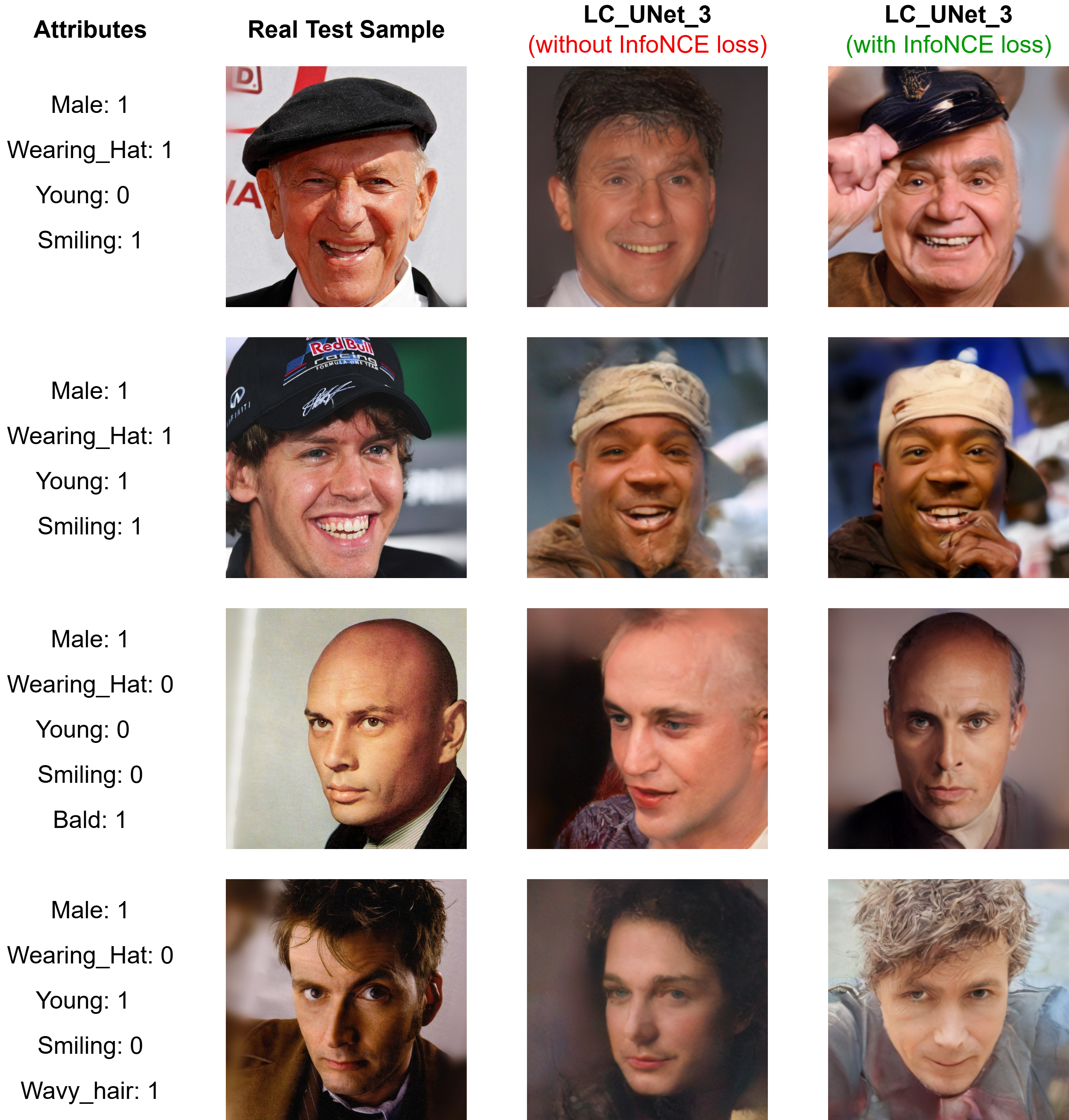}
\caption{Visual Comparison between LC\_UNet\_3 with and without InfoNCE loss training.}
\label{fig:infonce_viz}
\end{figure}

In Fig. \ref{fig:infonce_viz}, the "Without InfoNCE" images (middle) lack clarity, with blurred features, distorted proportions, and unnatural textures. The "With InfoNCE" images (right) are sharper, with better-defined facial features, more accurate attributes generation, and improved realism, though background integration and minor distortions remain issues in both sets. Overall, InfoNCE significantly enhances quality.

\begin{table}[h]
\centering
\caption{Comparison of Latent Conditional UNet architectures. The Up Blocks are symmetric to the Down Blocks, with cross-attention (e.g., CrossAttnUpBlock2D) and normal blocks (e.g., UpBlock2D) applied at corresponding resolution levels in the encoding and decoding paths.}
\begin{tabular}{lllll}
\toprule
\textbf{Model} & \textbf{LC\_UNet\_Base} & \textbf{LC\_UNet\_3} & \textbf{LC\_UNet\_5} & \textbf{LC\_UNet\_6} \\ \midrule
\textbf{Number of Blocks} & 6 & 5 & 6 & 6 \\
\textbf{Block Out Channels} & 128,128,256,256,512,512 & 128,128,256,512,512 & 128,128,256,256,512,512 & 128,128,256,256,512,512 \\
\textbf{Input Size} & 64$\times$64 & 64$\times$64 & 64$\times$64 & 64$\times$64 \\ \midrule
\textbf{Down Blocks} & DownBlock2D & DownBlock2D & DownBlock2D & CrossAttnDownBlock2D \\
\textbf{} & DownBlock2D & CrossAttnDownBlock2D & DownBlock2D & CrossAttnDownBlock2D \\
\textbf{} & DownBlock2D & DownBlock2D & DownBlock2D & DownBlock2D \\
\textbf{} & DownBlock2D & DownBlock2D & DownBlock2D & DownBlock2D \\
 & CrossAttnDownBlock2D & DownBlock2D & CrossAttnDownBlock2D & DownBlock2D \\
 & DownBlock2D &  & CrossAttnDownBlock2D & DownBlock2D \\ \midrule
\textbf{Parameters} & 140.45 M & 104.95 M & 168,04 M & 116,84 M \\ \bottomrule
\end{tabular}
\label{tab:unet_vertical_comparison}
\end{table}

\subsection{Combined Conditioning: Segmentation and Attributes}

To improve controllability and semantic precision in face generation, we developed a combined conditioning pipeline that leverages both facial attributes and segmentation masks. This approach enhances structural fidelity by encoding global attributes alongside spatial part-based information.

\textbf{Architecture.} The conditioning mechanism includes two parallel branches:

\begin{itemize}
    \item \textbf{Attribute Branch:} 40-dimensional multi-hot vectors are passed through an MLP to produce a 128-dimensional embedding. This embedding is trained using InfoNCE loss to ensure semantic clustering of similar attribute configurations.
    
    \item \textbf{Segmentation Branch:} Ten binary part-wise masks (e.g., for eyes, lips, hat, glasses) are stacked and processed by a SegFormer encoder. We evaluated two variants: (1) a pretrained SegFormer from Hugging Face \texttt{nvidia/segformer-b0-finetuned-ade-512-512}, and (2) a custom-trained SegFormer model fine-tuned on the CelebAMask-HQ segmentation masks. The encoder output is mean-pooled and projected into a 128-dimensional space via a linear layer.
\end{itemize}

The two 128-dimensional embeddings are concatenated and projected into a single 128-dimensional vector, which is used as the cross-attention input to the \texttt{UNet2DConditionModel} at each resolution level.

\textbf{Custom SegFormer Training.} To improve domain alignment, we trained a SegFormer model on the CelebAMask-HQ dataset. Training involved binary part-wise masks as input and cross-entropy loss over semantic regions. 

\textbf{Training Configuration.} We used a VQ-VAE to encode 256×256 RGB images into 64×64×4 latent tensors. The DDPM denoising model was trained on 2,700 samples using 1,000 diffusion steps, with EMA and mixed precision enabled. The segmentation and attribute paths were trained jointly using only image reconstruction loss; InfoNCE was applied to attributes only.

\textbf{Results.} Combined conditioning led to improved localization and fidelity in face synthesis. Visually, segmentation-enhanced models preserved facial boundaries and accessories more consistently. This dual-encoder approach demonstrated the complementary nature of global attributes and local part-awareness in conditioning latent diffusion models.

\subsection{LoRA Fine-tuning Setup for Stable Diffusion} \label{appendix:lora_stable_diffusion}

\begin{itemize}
    \item \textbf{Setup 1 (Custom LoRA Training)}: This version offers full control over training logic and model behavior. LoRA is applied manually to UNet attention layers using \texttt{get\_peft\_model}, and training is executed via a custom loop that integrates learning rate scheduling, Exponential Moving Average (EMA), FID evaluation, and conditional text prompt support. A notable feature of this pipeline is a \textbf{weighted timestep sampling strategy}, which dynamically emphasizes different parts of the diffusion trajectory across epochs—focusing on later timesteps early in training, mid-range steps in the middle phase, and early steps toward the end. This improves denoising precision and sample quality.

    \item \textbf{Setup 2 (Minimal LoRA Training)}: This implementation uses the same model backbone and LoRA configuration but streamlines the training loop by omitting advanced features like EMA and evaluation modules. While conditional training is supported via prompt-text conditioning, configurability during training is reduced compared to Setup 1.
\end{itemize}

\begin{table*}[ht]
\centering
\caption{Comparison of LoRA Fine-Tuning Pipelines for Stable Diffusion v2}
\label{tab:lora-v1-v2-comparison}
\begin{tabular}{p{4.2cm}p{5.8cm}p{5.8cm}}
\toprule
\textbf{Aspect} & \textbf{Setup 1 (Custom LoRA Training)} & \textbf{Setup 2 (Minimal LoRA Training)} \\
\midrule
\textbf{LoRA Integration} & Manual via \texttt{get\_peft\_model} and UNet injection & Same method, integrated with simplified training loop \\
\textbf{Pipeline Structure} & Fully custom: UNet + DDPM scheduler + EMA + FID evaluation & Lightweight setup using PEFT with minimal configurability \\
\textbf{Timestep Sampling} & Dynamic weighted sampling strategy across epochs & Fixed uniform timestep sampling \\
\textbf{Guidance Scale (CFG)} & 1.0 (unconditional default) & 7.5 (for stronger prompt alignment) \\
\textbf{Conditional Support} & Yes (with text prompts from CelebAMask-HQ) & Yes (same prompt mechanism) \\
\textbf{Unconditional Support} & Yes (using neutral placeholder prompts) & Yes \\
\textbf{Learning Rate} & $2 \times 10^{-4}$ & $1 \times 10^{-4}$ \\
\textbf{Dropout} & 0.0 & 0.1 \\
\textbf{EMA Updates} & Supported & Not included \\
\textbf{FID Evaluation} & Supported & Not included \\
\textbf{Overall Capability} & High flexibility and research extensibility & Simpler implementation with core training functionality \\
\bottomrule
\end{tabular}
\end{table*}

\begin{table}[h]
\centering
\caption{Full FID ($\downarrow$) scores on our test set for LoRA fine-tuned Stable Diffusion v2.}
\label{tab:lora-fid-appendix}
\begin{tabular}{lccc}
\toprule
\textbf{Inference Steps} & \textbf{Condition} & \textbf{Setup} & \textbf{FID} $\downarrow$ \\
\midrule
50   & Unconditional     & No Tuning & 297.098 \\
150  & Unconditional     & No Tuning & 315.0361 \\
50  & Conditional     & No Tuning & 114.729 \\
150  & Conditional     & No Tuning & 120.4432 \\
\midrule
50   & Conditional     & 1 & 67.515 \\
150  & Conditional     & 1 & 65.305 \\
50   & Unconditional   & 1 & 91.315 \\
150  & Unconditional   & 1 & 119.082 \\
50   & Conditional     & 2 & 70.912 \\
150  & Conditional     & 2 & 70.542 \\
50   & Unconditional   & 2 & 95.200 \\
150  & Unconditional   & 2 & 142.042 \\
\bottomrule
\end{tabular}
\end{table}

\begin{figure}[h]
\centering
\begin{tabular}{cc}
\includegraphics[width=0.14\linewidth]{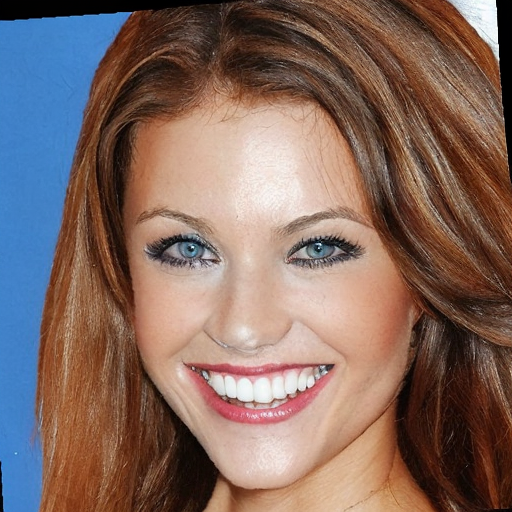} &
\includegraphics[width=0.14\linewidth]{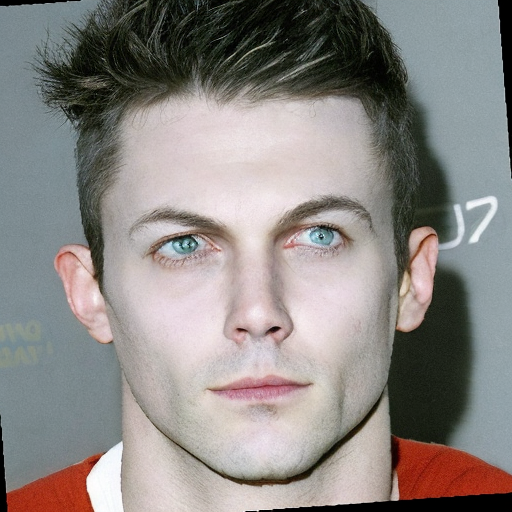} \\
\includegraphics[width=0.14\linewidth]{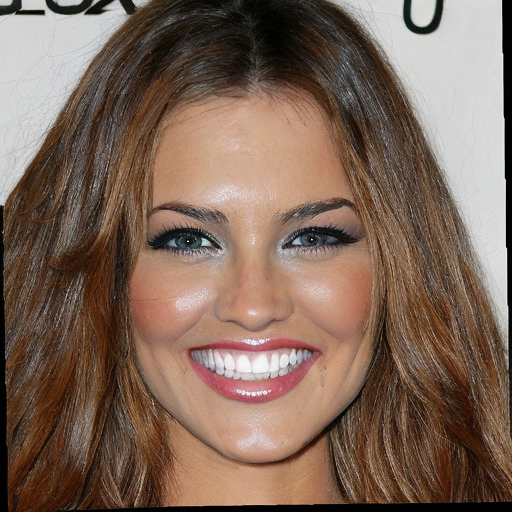} &
\includegraphics[width=0.14\linewidth]{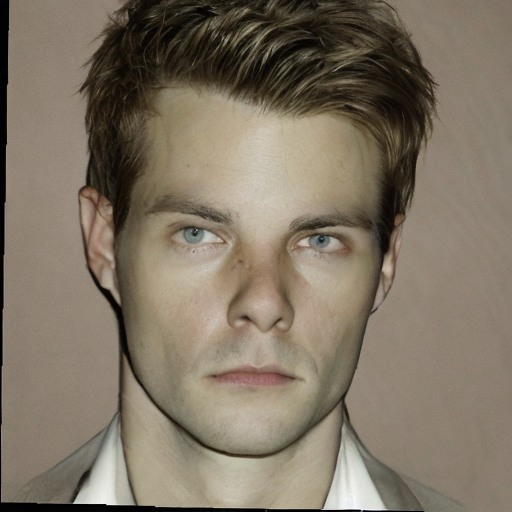} \\
\end{tabular}

\vspace{0.5em}
\begin{tabular}{p{0.42\linewidth} p{0.42\linewidth}}
\centering \textit{Prompt 1: A photo of a person with arched eyebrows, attractive, brown hair, heavy makeup, high cheekbones, mouth slightly open, no beard, oval face, smiling, wearing lipstick, young} & 
\centering \textit{Prompt 2: A photo of a person with 5 o clock shadow, arched eyebrows, male, no beard, pale skin, young} \\
\end{tabular}

\caption{Effect of inference steps on conditional LoRA generation using Stable Diffusion v2. Each column shows outputs generated from the same prompt at Inference Step 50 (top) and 150 (bottom) denoising steps. More steps generally yield sharper, more coherent results.}
\label{fig:inference_steps_lora}
\end{figure}

\begin{figure}[h]
\centering
\begin{tabular}{cccc}
\includegraphics[width=0.12\linewidth]{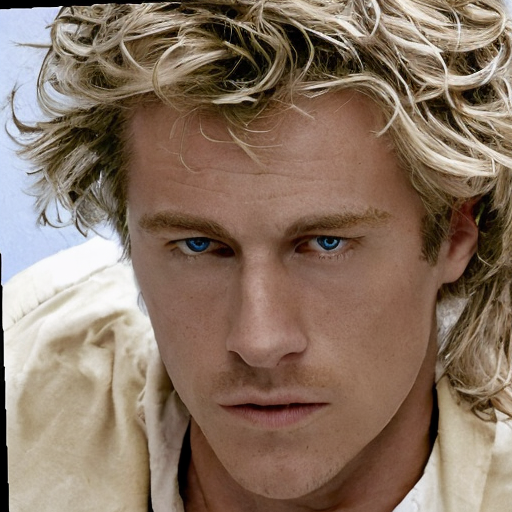} &
\includegraphics[width=0.12\linewidth]{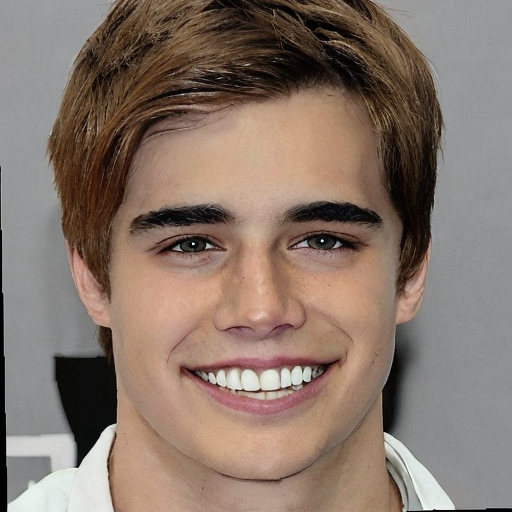} &
\includegraphics[width=0.12\linewidth]{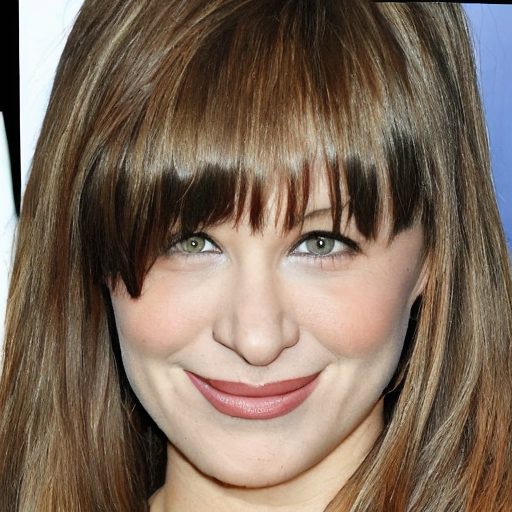} &
\includegraphics[width=0.12\linewidth]{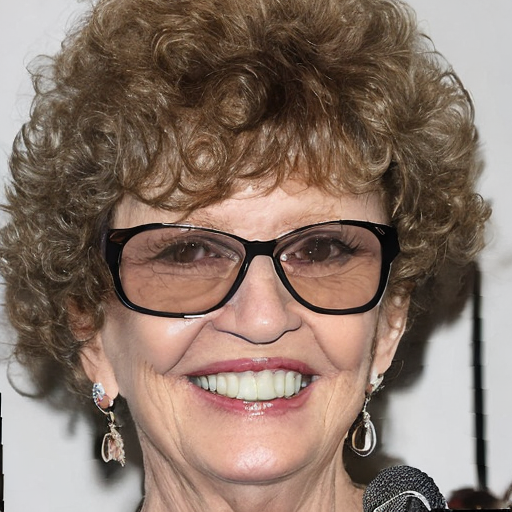} \\
\includegraphics[width=0.12\linewidth]{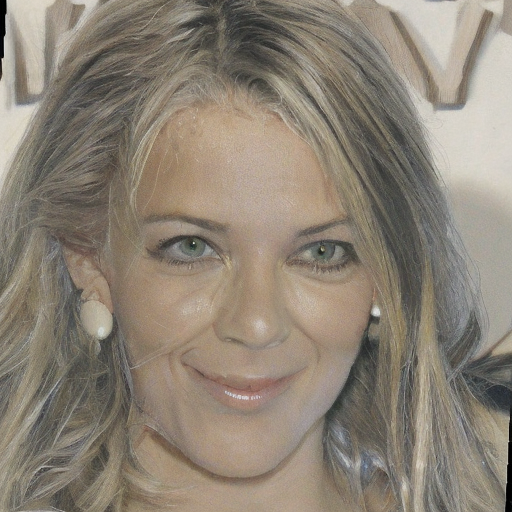} &
\includegraphics[width=0.12\linewidth]{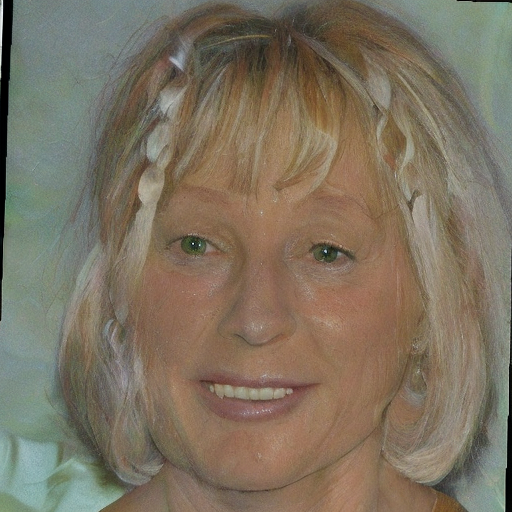} &
\includegraphics[width=0.12\linewidth]{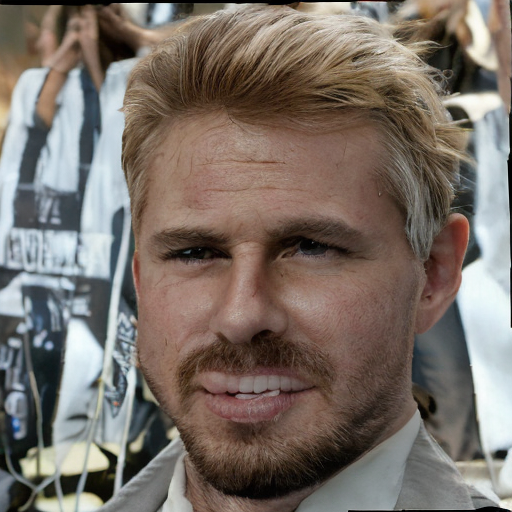} &
\includegraphics[width=0.12\linewidth]{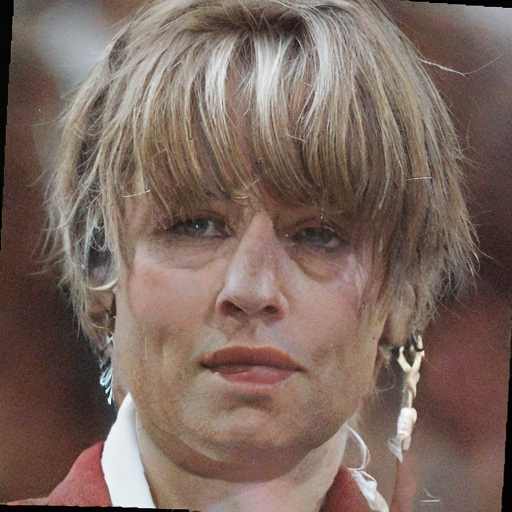} \\
\end{tabular}
\caption{Qualitative comparison of Conditional (top row) vs. Unconditional (bottom row) LoRA-generated samples using Stable Diffusion v2 (Version 1, 150 inference steps, 512×512 resolution). Conditional models produce more coherent facial structures, while unconditional outputs suffer from color shifts and reduced fidelity.}
\label{fig:lora_cond_vs_uncond}
\end{figure}

\begin{figure}[h]
\centering
\begin{tabular}{ccccc}
\includegraphics[width=0.12\linewidth]{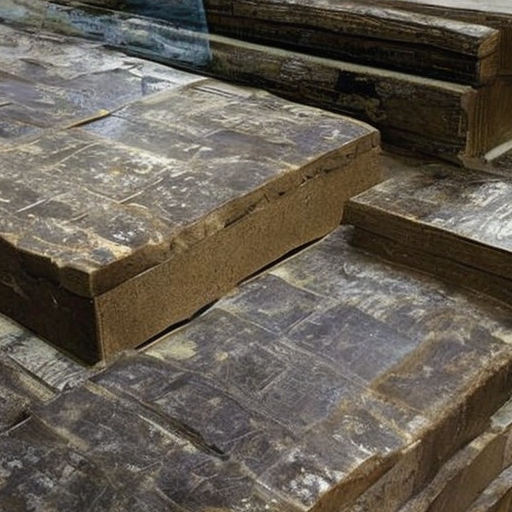} &
\includegraphics[width=0.12\linewidth]{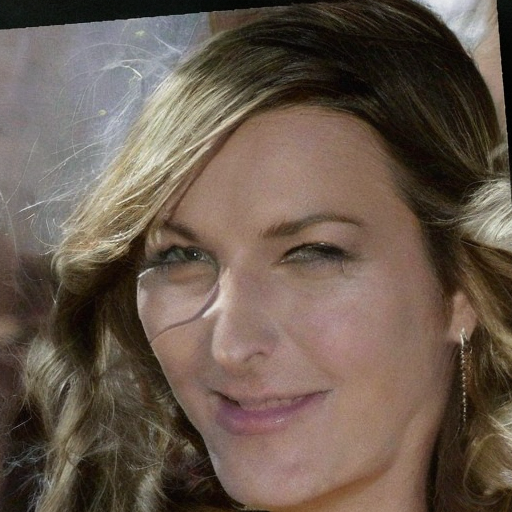} &
\includegraphics[width=0.12\linewidth]{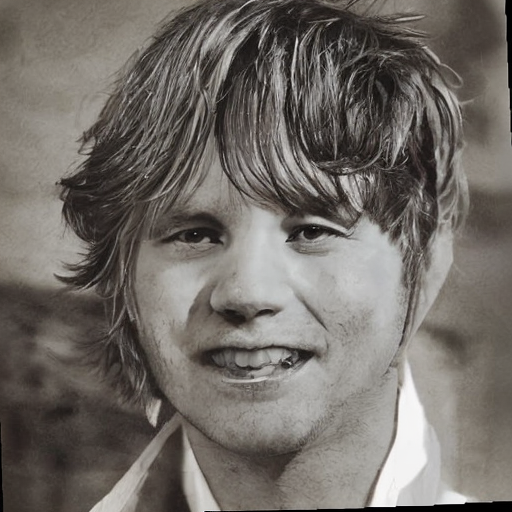} &
\includegraphics[width=0.12\linewidth]{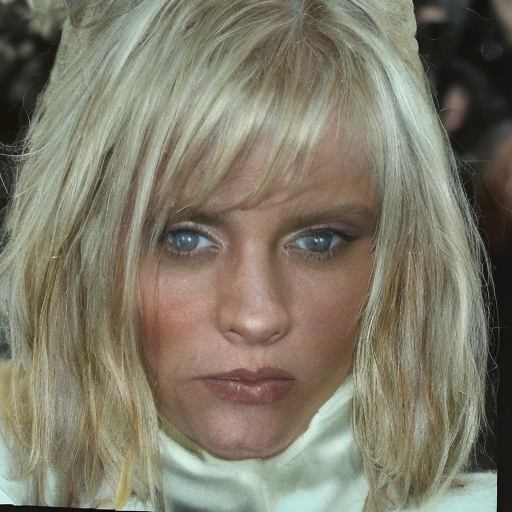} &
\includegraphics[width=0.12\linewidth]{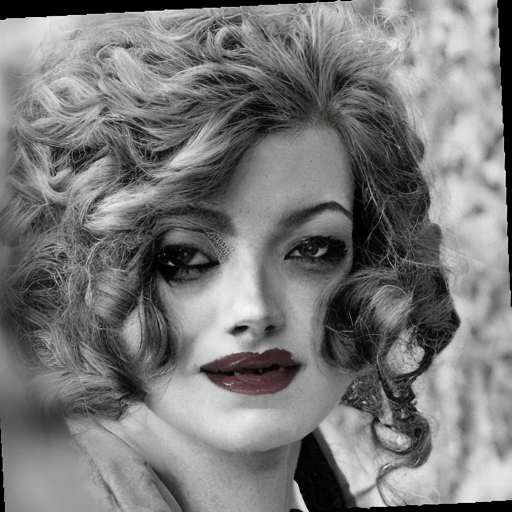} \\
Step 0 & Step 1 & Step 2 & Step 3 & Step 4 \\
\end{tabular}
\caption{Progression of unconditional LoRA-generated images over training steps (Stable Diffusion v2, 512×512 resolution). Outputs evolve from non-face to increasingly realistic faces.}
\label{fig:uncond_progression}
\end{figure}

\end{document}